\documentclass[sigconf]{acmart}

\AtBeginDocument{%
  \providecommand\BibTeX{{%
    \normalfont B\kern-0.5em{\scshape i\kern-0.25em b}\kern-0.8em\TeX}}}
    
\usepackage{algorithm}
\usepackage{algpseudocode}
\usepackage{enumitem}

\setcopyright{none}
\acmConference[]{PREPRINT}{July}{2023}
\acmDOI{10.48550/arXiv.2305.01264 }
\acmISBN{}

\newcommand\new[1]{#1} 

\newcommand\ourMethodLong{Multi-Task Multi-Behavior MAP-Elites}
\newcommand\ourMethodAccronyme{MTMB-MAP-Elites}

\begin{document}

\title{\ourMethodLong}

\author{Timothée Anne}
\orcid{0000-0002-4805-0213}
\authornotemark[1]
\author{Jean-Baptiste Mouret}
\orcid{0000-0002-2513-027X}
\authornotemark[1]
\email{firstname.name@inria.fr}
\affiliation{%
  \institution{Inria Nancy - Grand Est, Université de Lorraine, LORIA, CNRS}
  \city{Nancy}
  \country{France}
}

\renewcommand{\shortauthors}{Anne and Mouret}

\begin{abstract}
    We propose Multi-Task Multi-Behavior MAP-Elites, a variant of MAP-Elites that finds a large number of high-quality solutions for a large set of tasks (optimization problems from a given family). It combines the original MAP-Elites for the search for diversity and Multi-Task MAP-Elites for leveraging similarity between tasks. It performs better than three baselines on a humanoid fault-recovery set of tasks, solving more tasks and finding twice as many solutions per solved task.
\end{abstract}



\begin{teaserfigure}
  \includegraphics[width=\textwidth]{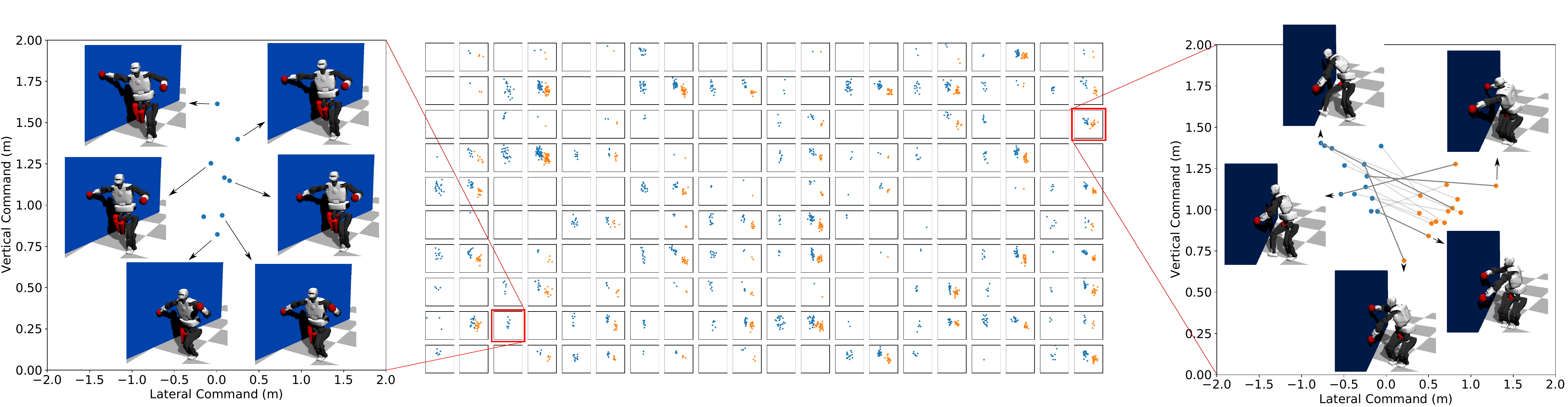}
  \caption{Archive collected over 200 tasks (100 situations with either the right-hand or with both hands), using \ourMethodAccronyme{} on a humanoid robot fault-recovery set of tasks. Each situation corresponds to the posture of the humanoid robot, a fault happening in its leg likely to make it fall, and the orientation and distance of a wall within arm's reach. The goal is to find as many different successful contact positions on the wall as possible. Each pair of boxes shows the collected successful commands for one situation, separated into two tasks: using only the right-hand (in blue) and using both hands (in blue and orange). We show different solutions and corresponding snapshots of the final posture of the robot using only the right-hand (on the left) and using both hands (on the right). }
  \label{fig:teaser}
\end{teaserfigure}

\maketitle

\section{Introduction}

Many problems involve searching for diverse solutions to a set of tasks. For instance, in grasping, one can be interested in finding various grasps for different objects to choose the most appropriate one when the object is partially obstructed (e.g., in a heap). Here, a \emph{task} corresponds to a specific object to be grasped.

On the one hand, MAP-Elites~\cite{MAP-Elites} creates an archive of diverse and high-performing solutions for a specific task (defined by the fitness function). On the other hand, Multi-Task MAP-Elites~\cite{Multi-Task_MAP-Elites} leverages the similarity between the tasks to find one high-quality solution for each task.

In this paper, we present a novel method called Multi-Task Multi-Behavior MAP-Elites (MTMB-MAP-Elites) that combines the features of both MAP-Elites~\cite{MAP-Elites} and Multi-Task MAP-Elites~\cite{Multi-Task_MAP-Elites}. Given a set of tasks, it builds an archive containing diverse solutions for each task.

In this particular work, we are interested in building an archive of reflexes for a humanoid robot. The goal is to use a wall to make the robot regain its balance after detecting a fault in its leg that has a high chance of making it fall. A task corresponds to discovering successful contact positions for a specific situation (posture of the robot, a fault occurring in its leg, and orientation of a wall within arm's reach). The command corresponds to the target contact positions (with one or two hands) which are set as high-weighted tasks in a whole-body controller. This is useful because humanoid robots are susceptible to faults due to their complex structure and bipedal stance. For instance, a single mechanical failure can make them fall~\cite{DRC_book, DRC_paper}. 

A previous work, D-Reflex~\cite{D-Reflex}, addresses this problem by building an archive using a grid search in simulation.
They only considered the use of the right-hand, while our approach aims to generalize to also using both hands on walls in any orientation around the robot. This prevents us from using a grid search which would require, using the same parameters, nearly $200,000$ simulations per situation due to the quadratic increase in the number of possible contact positions when using both hands. 

In this paper, we evaluate the performance of MTMB-MAP-Elites to find diverse solutions on the fault-recovery set of tasks using the Talos humanoid robot in simulation and compare it to three baselines: a random search, a grid search, and MAP-Elites~\cite{MAP-Elites} applied on each task.

\section{Problem formulation}
Given a set of tasks $T_{1:n}\in \mathcal{T}$, the goal is to find for each task $T_i$ as many different solutions $(s_i^j)_{j \le m_i}$ as possible where $m_i$ is the number of different solutions found for the task $T_i$. 

The task space $\mathcal{T}$ defines a command space $\mathcal{C}$, a fitness function $fitness: \mathcal{T} \times \mathcal{C} \rightarrow \mathbb{R}$, a behavior space $\mathcal{B}$, and a transition function $\mathcal{F}:\mathcal{T} \times \mathcal{C} \rightarrow \mathcal{B}$. We also hypothesize that the fitness function is bounded by $f_{max}\in \mathbb{R}$ so that we know if a command $c$ is optimal for a given task $T$. 

More formally, the problem formulation is the following:

\begin{equation}
    \begin{array}{l}
         \text{maximize} \sum_{i=1}^{n} m_i \\ 
         \text{ s.t. } \forall i, \forall j \le m_i, fitness(T_i, c_i^j) = f_{max} \\
         \text{ s.t. } \forall i, \forall k \ne j, \mathcal{F}(T_i, c_i^j) \ne \mathcal{F}(T_i, c_i^k) \\
    \end{array}
\end{equation}

\section{Algorithm}
To solve this problem we propose \ourMethodAccronyme{} a combination of MAP-Elites~\cite{MAP-Elites} and Multi-Task MAP-Elites \cite{Multi-Task_MAP-Elites}. The main difference is that we do not search for the best solution in each task but for the greatest number of diverse solutions in each task. We hypothesize that the different tasks have enough similarity between them to share some solutions which justifies solving them together to save time and computation compared to using MAP-Elites~\cite{MAP-Elites} for each task individually. 

For clarity, in the remainder of the paper we call elite a command $c$ that has been evaluated and stored in the archive, and solution an optimal elite, e.g., an elite with maximal fitness $f_{max}$. 

The algorithm is the following:

\begin{enumerate}
    \item Initialization:
    \begin{enumerate}
         \item select a budget $B$ of evaluations (depending on the available time and computational resources);
         \item select $n$ random tasks $T_{1:n}\in \mathcal{T}$ ($n$ should be largely inferior to $B$);
         \item initialize the archive using random commands on randomly selected tasks until we got enough elites (as a rule of thumb we choose to stop when the algorithm has found $n$ elites);
    \end{enumerate}
   \item Core algorithm repeated $B-b_{init}$ times where $b_{init}$ is the number of initial evaluations:
    \begin{enumerate}
        \item select the command $c$ by randomly picking two tasks $T_i$ and $T_j$ with elites, select at random an elite from each one $c_i$ and $c_j$, and perform traditional cross-over and mutation operators; 
        \item select the task $T_k$ at random;
        \item evaluate the command $c$ by collecting the behavior $b_k = \mathcal{F}(T_k, c)$ and the corresponding fitness $f=fitness(T_k, c)$;
        \item update the archive: if the behavior $b_k$ was not present we add it to the task $T_k$, and if it was already present and the new fitness $f$ is greater than the previous one, we replace the old elite with the new one.
    \end{enumerate}
\end{enumerate}


\section{Experiment}

We evaluate \ourMethodAccronyme{} on a set of fault-recovery tasks for the humanoid robot Talos in simulation. The humanoid robot detects an unknown fault occurring in its leg (a combination of amputated, passive, or locked joints) which will most often induce a fall. It knows (using its onboard sensors) that there is a wall within arm's reach as well as its orientation and distance. The goal is to find successful contact positions on the wall using either the right-hand or both hands. 

\paragraph{Task Space}

We define the task space $\mathcal{T}$ as the set of different situations corresponding to the current posture of the robot, the configuration of the wall (distance and orientation to the robot), and the fault occurring in the robot's leg. 

To sample the tasks $T_{1:n}$, we randomly select:
\begin{itemize}
    \item target Cartesian positions for the left and right hands (which after 4s will put the robot in a reachable posture); 
    \item configurations of the wall (distance and orientation);
    \item faults (a combination of passive, locked, or amputated joints on the six joints of the right leg).
\end{itemize}

We preemptively run the situations in simulation, reject those where the robot makes contact with the wall before the fault occurs, and sample new ones. 

\paragraph{Command Space}
The command corresponds to the \emph{target} contact positions on the wall. The robot is controlled using a whole-body controller~\cite{dalin_whole-body_2021} which solves at high frequency (e.g., 500Hz) a quadratic programming optimization representing the set of tasks and constraints using the model of the undamaged robot. Our \emph{target} contact positions are set as high-weighted tasks in the optimization and can be seen as high-level commands. 

We explicitly promote diversity by duplicating each situation in two: one task for using the right-hand and one task for using both hands. We set $\mathcal{C} = X\times Z\times X\times Z$ where $X=[x_{min}, x_{max}]$ and $Z=[z_{min}, z_{max}]$ are set to broadly represent reachable contact positions on the wall. For the tasks using only the right-hand, only the first two dimensions are relevant. 

\paragraph{Behavior Space}
The behavior space $\mathcal{B}$ is defined as the Cartesian \emph{reached} contact positions of the hands on the wall (in 2D when using right-hand and in 4D when using both hands). We make a difference between the \emph{target} contact position and the \emph{reached} contact position because we found out empirically that due to the mismatch between the known undamaged model of the robot and the current damaged robot, the reached contact position can be far (e.g., several centimeters) from the target contact position. 

To differentiate behaviors, we set squares of side 20cm and pick as behavior $b = \mathcal{F}(T, c)$ the index of the square (or squares when both hands are used) containing the \emph{reached} contact positions. As the set of possible behaviors is large and a lot of them are not realizable (for reachable or auto-collision reasons for example) we initialize an empty map and append new behaviors as they are discovered.  

\paragraph{Fitness Function}
The fitness corresponds to the time before the simulation stops due to an auto-collision, a fall (unplanned contact between the robot and the floor or the wall), or the timeout of 10s ($f_{max}=10s$). This timeout has been selected as a good threshold between reducing simulation time and reducing the number of unstable solutions (when the robot does not fall before the timeout but would have fallen after); an early experiment has shown that by increasing the timeout to 15s, we only detect one unstable solution among 310 solutions.

\begin{figure*}[ht]
    \centering
    \includegraphics[width=\linewidth]{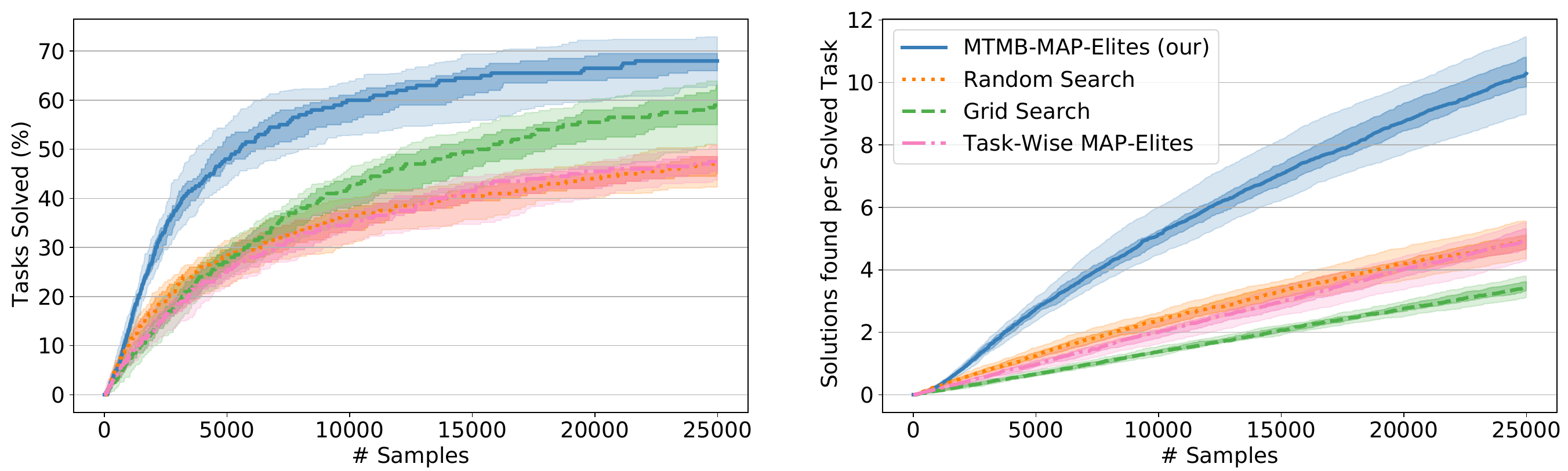}
    \caption{Comparison for solving 200 tasks (100 situations with either the right-hand or with both hands) between \ourMethodAccronyme{} and three baselines: a random search, a grid search, and MAP-Elites on each task individually. (a) The percentage of tasks solved and (b) the number of solutions per solved task. The line represents the median, the darker shaded area the [25\%, 75\%] quantiles of the data, and the lighter shaded area [5\%, 95\%] quantiles on 25 replications. \ourMethodAccronyme{} solves more tasks than the best baselines ($+20.8\%$ than Random Search, $+9.9\%$ than Grid-Search, and $+20.7\%$ than Task-Wise MAP-Elites) and more importantly finds more than two times as many solutions per solved task. }
    \label{fig:main}
\end{figure*}

\subsection{Evaluation}

We sample 100 situations and duplicate each one in two tasks (the right-hand and both hands), which gives us $n=200$ tasks. We set a budget of $B=25,000$ simulations, stopping the initialization when the archive had gathered 200 elites.

To our knowledge, there exists no method in the literature to solve this particular problem. As we want a diverse set of solutions we cannot compare to black-box optimizations such as CMA-ES\cite{CMA-ES}.

We compare against three naive baselines with the same evaluations budget $B=25,000$:
\begin{itemize}
    \item Random Search, e.g., selecting the command using a uniform distribution in $\mathcal{C}$.
    \item Grid Search, e.g., always selecting the same commands evenly panning the command space $\mathcal{C}$ for each task (to have the same number of evaluations per task (e.g., $\tfrac{B}{n}=125$) we use a $5\times5$ grid for the right-hand and a $5\times2$ grid for both hands); 
    \item Task-Wise MAP-Elites, e.g., running MAP-Elites~\cite{MAP-Elites} on each task with a budget of 125 simulations.
\end{itemize}

\new{Grid-search and Task-Wise MAP-Elites are inherently sequential on each task. To facilitate a more accurate comparison with our approach, we have randomized the sequence in which the various tasks are evaluated.}

We performed 25 replications and evaluate: 
\begin{itemize}
    \item the percentage of solved tasks (e.g., tasks with at least one solution);
    \item the number of solutions per solved task. 
\end{itemize}


\subsection{Result}
Figure~\ref{fig:main} presents the evaluation of \ourMethodAccronyme{} against the three baselines over 25 replications. Figure~\ref{fig:teaser} presents an example of a collected archive using \ourMethodAccronyme{} with snapshots of the final postures of the robot for the solutions of two tasks. 


\ourMethodAccronyme{} outperforms all baselines, solving in average $67.8\% \pm 3.7\%$ tasks against 
$47.0\% \pm 2.8\%$ for Random Search,  
$57.9\% \pm 4.3\%$ for Grid Search, 
and $47.1\% \pm 2.6\%$ for Task-Wise MAP-Elites. 

More importantly for our goal to build a diverse dataset, \ourMethodAccronyme{} finds an average $10.2 \pm 0.8$ solutions per solved task against 
$4.9 \pm 0.4$ for Random Search, 
$3.4 \pm 0.3$ for Grid Search, 
$4.9 \pm 0.4$ for Task-Wise MAP-Elites.

Our hypothesis is that with a budget of 125 evaluations per task, Task-Wise MAP-Elites does not have the time to become efficient. In the first step, it only performs a random search and then has to quickly exploit the few elites it has founds leading to early exploitation and performances similar to using a random search.

\new{Grid Search covers the command space uniformly, enabling it to tackle almost as many tasks as our method. However, this approach also considerably diminishes its ability to uncover diverse solutions.} By merging commands from two different elites, MTMB-MAP-Elites indirectly samples from a distribution that better represents the subspace of possible solutions, thus saving time by discarding regions where there are low chances to discover a solution.

\section{Conclusion}
Our method, \ourMethodLong{}, demonstrates superior performance over three baselines, solving significantly more tasks ($+20.8\%$ than Random Search, $+9.9\%$ than Grid-Search, and $+20.7\%$ than Task-Wise MAP-Elites) and finding twice as many solutions per solved task on a set of fault-recovery tasks using a humanoid robot in simulation. 

In the next step, we will leverage the capabilities of \ourMethodLong{} to construct a dataset of diverse solutions, using privileged knowledge (the nature of the fault). This dataset will then be used to train a machine learning-based policy for selecting robust contact positions with one or two hands to prevent falling. 

This presents an alternative to Reinforcement Learning~\cite{SAC, PPO} that often lacks sufficient exploration. Initial experiments on our fault-recovery set of tasks show that PPO~\cite{PPO} often falls down to using only one hand because using both hands requires overcoming the deceptive low fitness of the self-collision between the two arms. 

\begin{acks}
Experiments presented in this paper were carried out using the Grid’5000 testbed, supported by a scientific interest group hosted by Inria and including CNRS, RENATER, and several universities as well as other organizations (see https://www.grid5000.fr). This project is supported by the CPER SCIARAT, the CPER CyberEntreprise, the Direction General de l’Armement (convention Inria-DGA ``humanoïde résilient''), the Creativ’Lab platform of Inria/LORIA, and the EurROBIN Horizon project (grant number 101070596).
\end{acks}

\bibliographystyle{ACM-Reference-Format}
\bibliography{biblio.bib}




\end{document}